\documentclass[letterpaper, 10 pt, conference]{ieeeconf}  % Comment this line out if you need a4paper

\IEEEoverridecommandlockouts                              % This command is only needed if 
\usepackage{graphicx}
\usepackage{comment}
\usepackage{amsmath,amssymb}
\usepackage{color}
\usepackage{url}
\usepackage{hyperref}
\usepackage{multirow}
\usepackage{booktabs}
\usepackage{bm}
\usepackage{siunitx}
\usepackage{todonotes}
\usepackage{tablefootnote}

\usepackage[acronym]{glossaries}

\newacronym{hsi}{HSI}{hyperspectral imaging}
\newacronym{hs}{HS}{hyperspectral}
\newacronym{cnn}{CNN}{Convolutional Neural Network}
\newacronym{ml}{ML}{machine learning}
\newacronym{bn}{BN}{batch normalization}
\newacronym{dr}{DR}{dimensionality reduction}
\newacronym{cw}{CW}{class weighting}
\newacronym{do}{DO}{dropout}
\newacronym{da}{DA}{data augmentation}
\newacronym{wd}{WD}{weight decay}
\newacronym{usm}{USM}{unsharp masking}
\newacronym{pca}{PCA}{principal component analysis}
\newacronym{pc}{PC}{principal component}
\newacronym{hcv}{HCV2}{HyperspectralCity V2.0}
\newacronym{ins}{INS}{inverse number of samples}
\newacronym{isns}{ISNS}{inverse square root of number of samples}
\newacronym{sgd}{SGD}{Stochastic Gradient Descent}
\newacronym{iou}{IoU}{intersection over union}
\newacronym{dl3}{DL3+}{DeeplabV3+}
\newacronym{prgb}{pRGB}{pseudo-RGB}
\newacronym{aspp}{ASPP}{atrous spatial pyramid pooling}

\newcommand{\hsiroad}{HSI-Road}
\newcommand{\hsidrive}{HSI-Drive}
\newcommand{\hyko}{HyKo2}
\newcommand{\hykovis}{\hyko-VIS}
\newcommand{\micro}{\ensuremath{\mu}}
\newcommand{\macro}{M}
\newcommand{\fone}{\textnormal{F}_1}

\newcommand{\foneM}{\textnormal{F}_{1_\macro}}

\newcommand{\acc}{\textnormal{Acc}}
\newcommand{\jaccard}{\textnormal{J}}

\newcommand{\eg}{e.\,g.}
\newcommand{\ie}{i.\,e.}

\newcommand{\runet}{RU-Net}

\glsdisablehyper

\title{\LARGE \bf
HS3-Bench: A Benchmark and Strong Baseline for Hyperspectral Semantic Segmentation in Driving Scenarios
}

\author{Nick Theisen, Robin Bartsch, Dietrich Paulus and Peer Neubert% <-this % stops a space
\thanks{This paper was accepted for IROS 2024 --- © 2024 IEEE.  Personal use of this material is permitted.  Permission from IEEE must be obtained for all other uses, in any current or future media, including reprinting/republishing this material for advertising or promotional purposes, creating new collective works, for resale or redistribution to servers or lists, or reuse of any copyrighted component of this work in other works. --- All authors are with the Institute of Computational Visualistics, University of Koblenz, Germany. Correspondence Email: {\tt\small nicktheisen@uni-koblenz.de}}%
}

\begin{document}

\maketitle
\thispagestyle{empty}
\pagestyle{empty}

%%%%%%%%%%%%%%%%%%%%%%%%%%%%%%%%%%%%%%%%%%%%%%%%%%%%%%%%%%%%%%%%%%%%%%%%%%%%%%%%
\begin{abstract}
Semantic segmentation is an essential step for many vision applications in order to understand a scene and the objects within. Recent progress in hyperspectral imaging technology enables the application in driving scenarios and the hope is that the devices perceptive abilities provide an advantage over RGB-cameras.
Even though some datasets exist, there is no standard benchmark available to systematically measure progress on this task and evaluate the benefit of hyperspectral data. 
In this paper, we work towards closing this gap by providing the HyperSpectral Semantic Segmentation benchmark (HS3-Bench). It combines annotated hyperspectral images from three driving scenario datasets and provides standardized metrics, implementations, and evaluation protocols. 
We use the benchmark to derive two strong baseline models that surpass the previous state-of-the-art performances with and without pre-training on the individual datasets.
Further, our results indicate that the existing learning-based methods benefit more from leveraging additional RGB training data than from leveraging the additional hyperspectral channels. This poses important questions for future research on hyperspectral imaging for semantic segmentation in driving scenarios. Code to run the benchmark and the strong baseline approaches are available under \url{https://github.com/nickstheisen/hyperseg}.

\end{abstract}

\section{INTRODUCTION}

Semantic segmentation models assign class labels to pixels and partition the image into regions with a problem-dependent semantic meaning. This is an important step in many vision applications such as scene understanding or the identification of image areas with certain task relevant properties.
\Gls{hsi} systems capture light in up to hundreds of very narrow spectral bands, often including ranges of the electromagnetic spectrum that are invisible to classical RGB-cameras and the human eye.
These advantages were exploited in the past to solve problems in many different domains, \eg ~remote sensing \cite{chenShallowGuidedTransformerSemantic2023}, medicine \cite{Calin2014HII} and agriculture \cite{Lu2020RAO}. 
The application in dynamic scenes is difficult, because many \gls{hsi}-sensors rely on scanning along the spectral dimension or along the spatial dimensions to capture a full hyperspectral cube, which takes time. 
However, through steady improvement of imaging systems, sensors became smaller, cheaper and simpler in usage. 
The latter is especially true for snapshot hyperspectral cameras which do not rely on scanning techniques but instead capture a full hyperspectral cube in an instant.
This progress enables the adoption of \gls{hsi}-sensors in novel applications and domains, \eg~dynamic driving scenarios, thus increasing the need for well-generalizing models for tasks such as semantic segmentation. However, the question whether \gls{hsi}-sensors provide a significant advantage in dynamic driving scenarios remains open.
\begin{figure}
    \centering
    \includegraphics[width=\columnwidth]{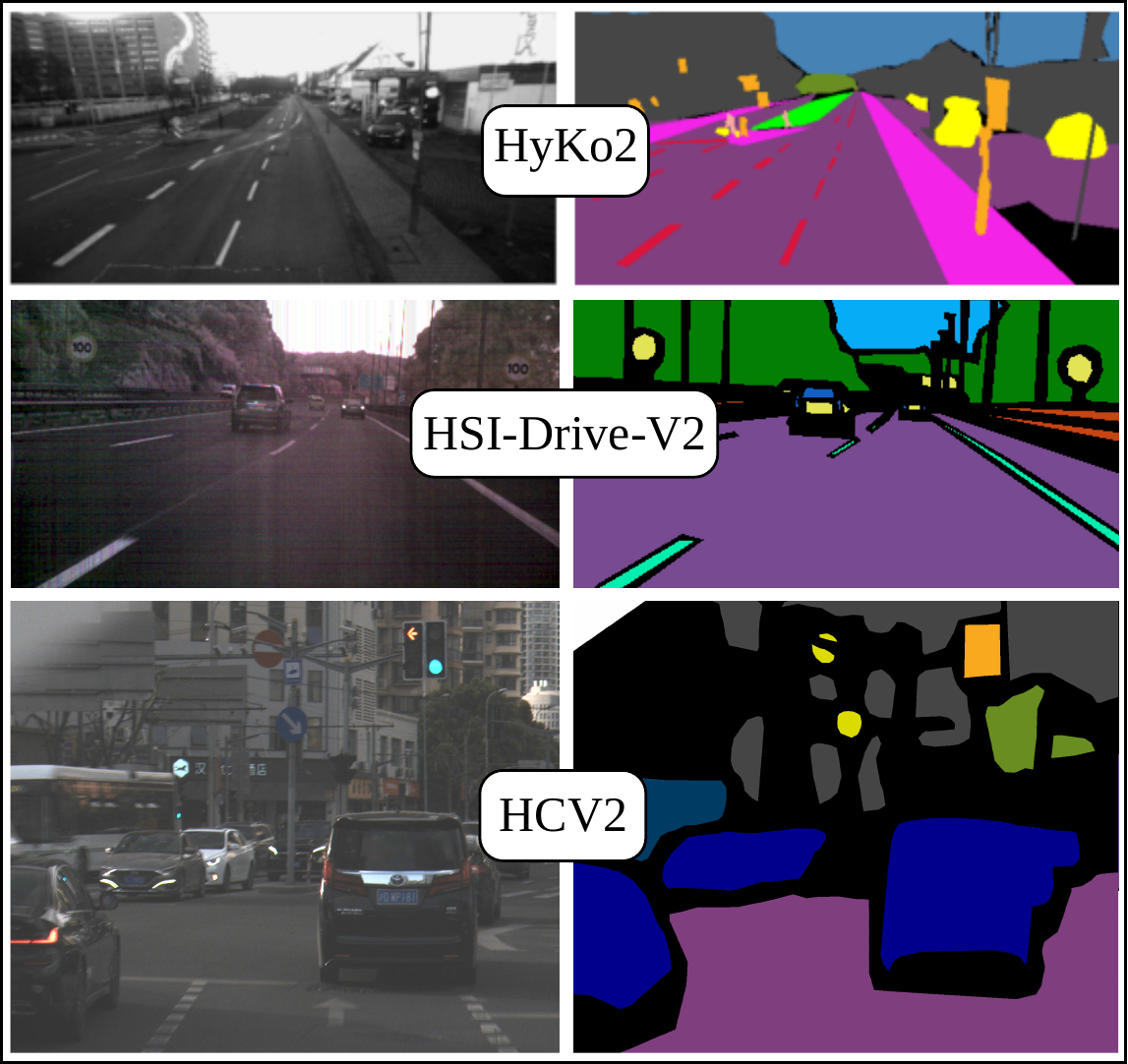}
    \caption{Examples scenes from the hyperspectral image datasets used in HS3-Bench (\hyko \cite{Winkens2017Hyko}, \hsidrive \cite{Basterretxea2021HSIDrive}, and \gls{hcv} \cite{Li2022HCV}) together with ground-truth semantic segmentation labels.}\vspace{-0.3cm}
    \label{fig:example_images}
\end{figure}

The lack of a standardized benchmark makes the comparison of different approaches very challenging. In the literature only evaluation results on individual datasets without a common evaluation protocol are published.  To address this problem, we present HS3-Bench, a hyperspectral semantic segmentation benchmark with the focus on driving scenarios. Example images can be seen in Fig.~\ref{fig:example_images}. The benchmark includes three datasets and allows systematic comparison of different approaches.

Our contributions can be summarized as follows:
\begin{itemize}
    \item We introduce HS3-Bench a hyperspectral semantic segmentation benchmark focused on driving scenarios systematic evaluation. 
    \item We propose two strong baselines for our benchmark, one based on the U-Net architecture, that uses only the \gls{hsi} data from the benchmark datasets and one based on \gls{dl3} that leverages additional data through pre-trained model weights. Our baselines outperform current state-of-the-art models.
    \item We provide evidence that the existing learning based-methods benefit more from leveraging additional RGB training data than from leveraging the additional \gls{hsi} channels. This poses important questions for future \gls{hsi} research.
\end{itemize}

This paper is organized as follows. Section~\ref{sec:relwork} discusses the availability of hyperspectral benchmarks in the literature. In \ref{sec:benchmark} we introduce HS3-Bench and present the used datasets, evaluation metrics as well as benchmark guidelines. Section~\ref{sec:baseline} describes the strong baselines. Description and results of our experiments can be found in section~\ref{sec:experiments}, followed by a summary of our findings in section~\ref{sec:summary}.

\section{Related Work} \label{sec:relwork}

Deep learning and particularly \gls{cnn}-based architectures establish the state of the art in semantic segmentation.
There exist multiple datasets with well-defined benchmarks for \textit{RGB-images} in driving scenarios, \eg ~Cityscapes Pixel-Level Semantic Labeling Task \cite{Cordts2016TCD} or KITTI Semantic Segmentation benchmark \cite{Alhaija2018IJCV}.
However, the combination of deep learning and \textit{\gls{hsi}} suffers from the limited amount of available data. Typically, hyperspectral datasets are small and therefore only very limited data is available for training. The most common datasets are remote sensing datasets consisting only of a single image, which can be considered solved, \eg\ \cite{chenShallowGuidedTransformerSemantic2023}. The train-test split is created by splitting the image into pixels (or patches) which restricts the applicable models to pixel classification models. 
In recent years, some larger, multi-image datasets showing urban and driving scenarios have been published (\hyko\ \cite{Winkens2017Hyko}, HSI-Drive v1 \cite{Basterretxea2021HSIDrive} and v2 \cite{Zaballa23HSIDrive2}, \gls{hcv} \cite{Li2022HCV}, \hsiroad\ \cite{Lu2020HSIRoad}). They allow the application of encoder-decoder models that predict pixel-precise classification maps for whole images during inference, such as U-Net \cite{Ronneberger2015UNet} or \gls{dl3} \cite{Chen18DeeplabV3Plus}.
Unfortunately, for \hyko\ a well-defined benchmark does not exist and \hsiroad\ consists of only two classes -- road and not road -- from which only one is considered during evaluation, making it closer to segmentation than semantic segmentation. Results for \hsidrive\ were published in \cite{Zaballa23OnChip} and  \cite{Zaballa23HSIDrive2} using a fully-convolutional network but in their experiments the authors use at most six different classes, as they combine certain combinations of minority classes into the class 'other'. On \gls{hcv} the authors of \cite{Ding23DualFusion} achieved good results with a dual stream model using \gls{hsi} as well as synthesized RGB. This allowed them to use a ResNet50 \cite{He16Resnet} backbone pre-trained on ImageNet, which led to a significant improvement of the model performance.

\iffalse
Even though the above mentioned multi-image datasets exist, they are still small compared to RGB datasets (\eg\ CityScapes \cite{Cordts2016TCD} or MS COCO \cite{Lin2014MCC}). In combination with the high dimensionality of hyperspectral data this poses a significant risk for models to overfit or suffer from the curse of dimensionality \cite{Hughes1968OTM}.
%Another problem is that of training robust and well-generalizing models. 
 A common way to mitigate such risks is through model regularization.
However, there exists a large variety of different regularization strategies each of which must be parameterized carefully, otherwise they may negatively impact the models' performance. Individual regularization approaches have been used with small hyperspectral datasets, \eg\ data augmentation \cite{Li2018Data,Nalepa2019Training}, batch normalization and dropout \cite{Abasi2019CNN}. Systematic investigation of the effectiveness of regularization techniques does not yet exists in this context, therefore we examine the effect of five common and well-understood regularization strategies (\gls{da}, \gls{wd}, \gls{cw}, \gls{bn} and \gls{do}) on a U-Net's performance and derive a regularized U-Net as a strong baseline.
\fi

\section{Benchmark: HS3-Bench}\label{sec:benchmark}

This section describes HS3-Bench, the HyperSpectral Semantic Segmentation Benchmark for driving scenarios. Detailed results and implementations are available in the benchmark repository.

\subsection{Datasets}

\begin{table}
    \footnotesize

    \caption{Overview of the \gls{hsi} datasets used in HS3-Bench}
    \centering
    \resizebox{\columnwidth}{!}{%
    \begin{tabular}{c|ccc}
        Name & \hykovis & \gls{hcv} & \hsidrive \\\hline
        Image size & $254 \times 510$ & $1400 \times 1800$ & $409 \times 216$ \\
        Bands & $15$ & $128$ & $25$ \\
        Range (nm) & $470$-$630$ & $450$-$950$ & $600$-$975$ \\
        Images & $371$ & $1330$ & $752$ \\
        Classes & $10$ & $19$ & $9$ \\
        Train/Test/Val-split ($\%$) & $50$/$20$/$30$ & $72$/$8$/$20$ & $60$/$20$/$20$\\
    \end{tabular}\vspace{-0.2cm}
    }
        \label{tab:used_datasets}
\end{table}
For our benchmark we built upon the three existing datasets 
\hyko\ \cite{Winkens2017Hyko},
\hsidrive\ \cite{Basterretxea2021HSIDrive}, and
\gls{hcv} \cite{Li2022HCV}. 
Example images can be seen in Fig.~\ref{fig:example_images}.
An overview over the datasets is given in \tablename~\ref{tab:used_datasets}.
We do not include the \hsiroad\ \cite{Lu2020HSIRoad} dataset since it only uses two classes (cf. section~\ref{sec:relwork}). All datasets show a high class imbalance, as apparent in many semantic segmentation datasets.

\textbf{\hykovis} consists of images showing urban and rural driving scenarios segmented into 10 classes with classes "road" (35.8\%), "sky" (15.2\%), "grass" (14.7\%) and "vegetation" (14.1\%) being most apparent, and "lane markers" (1.1\%), "panels" (1.5\%) and "person" (.03\%) being least apparent. As the name suggests the images are captured in the visual light spectrum.
 We omitted the near-infrared data set of \hyko\ because of the unfavorable sensor setup, which resulted in most images only showing a small patch of street in front of the vehicle.

\begin{figure}[t]
    \centering
    \includegraphics[width=\columnwidth]{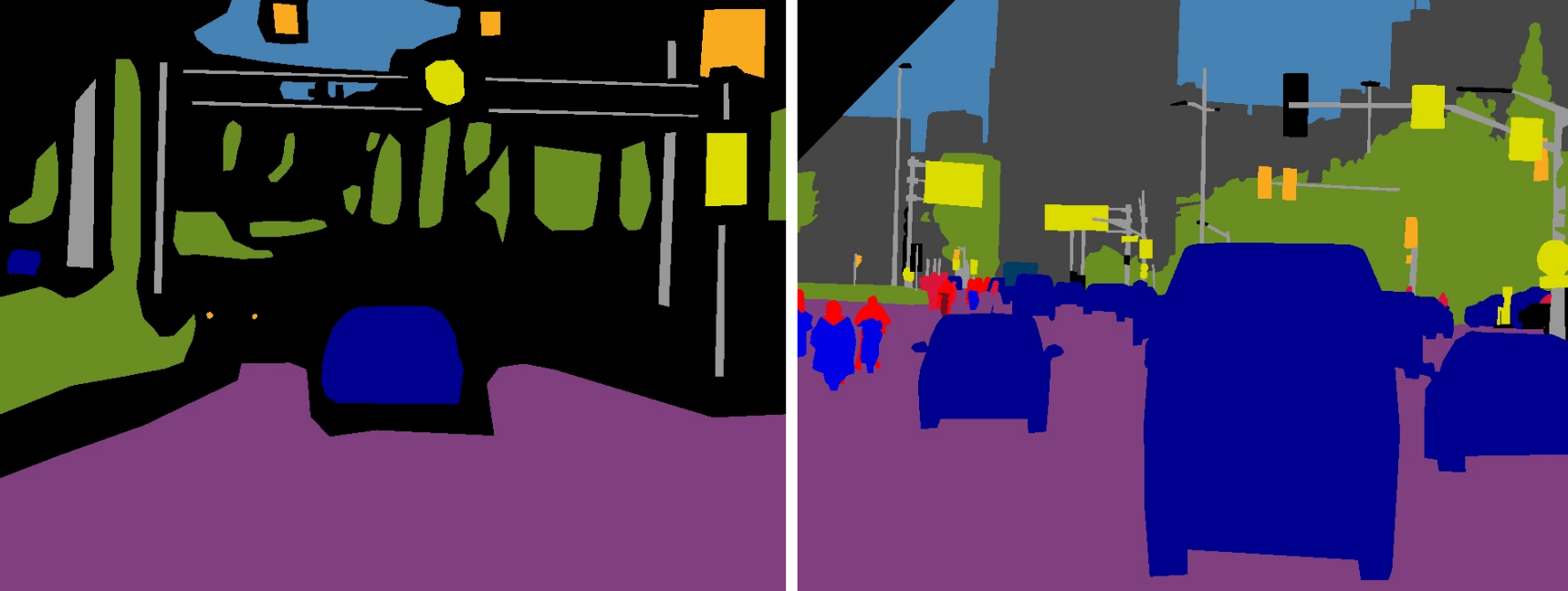}
    \caption{Coarse labels from \gls{hcv} train set (left) and fine labels from test set (right).}\vspace{-0.3cm}
    \label{fig:hcv-labels}
\end{figure}

\textbf{\gls{hcv}} was published in the context of the 2021 physics based learning ICCV workshop and shows urban scenarios. It has the highest spectral and spatial resolution and spectral range of the available datasets, covering the visual and near-infrared spectrum. With 19 classes it also has the highest number of distinct class labels. However, eight of the available classes make up less than 1\% of all labeled data and there exist no samples for the class "pole". The train set includes only coarse labels, while the test set includes fine labels, \ie\ objects are very precisely labeled as shown in Fig. \ref{fig:hcv-labels}.

\textbf{\hsidrive} contains 10 classes with "road" (60.7\%) and "vegetation" (21.3\%) being most apparent. All other classes are below 6\%, the authors even propose to not use the class "water" (.03\%) because of the low amount of samples available. We follow this recommendation in our benchmark and ignore pixels with this class. When we mention \hsidrive\ in this paper we always refer to v2 if not stated otherwise, as it includes v1.

\gls{hcv} comes with a defined train-test split. For the others we calculate dataset splits using pytorch's \textit{random\_split}-method to sample train-val-test-splits with proportions as shown in \tablename~\ref{tab:used_datasets}.
The sample lists for each split and each dataset will be published with the code.

% Further, we used a fixed seed ($=42$) to initialize network parameters for reproducibility as well as to generate train-test-splits for the datasets\footnote{We used pytorch \textit{random\_split}-method and exported the image-names to sample lists. The sample lists for each split and each dataset will be published with the code}. 

\subsection{Metrics}
The benchmark implements the commonly used metrics Accuracy ($\acc$), F1-Score ($\fone$) as defined in \cite{Sokolova2009Systematic}, and Jaccard Index ($\jaccard$) as defined in \cite{Everingham2015PascalVOC}.

\textbf{Micro- and macro-averaging~\cite{Sokolova2009Systematic}:}
Accuracy is calculated in micro- and macro-averaged form, denoted with $\micro$ and $\macro$ in subscript, respectively. All other metrics are only calculated in macro averaged form.
 With micro averaging, the metric is calculated by averaging over all sample predictions (per pixel), while in macro averaging the metric is calculated per class and the average is then calculated over all class-wise scores. As we are dealing with highly class imbalanced datasets, our evaluation is mostly focused on macro averaged metrics, as micro averaged metrics can be very much influenced by a data set's majority class. For example in an extreme case of a binary classification problem where $99\%$ of pixels belong to class A a classifier that always predicts A will have a micro-averaged accuracy of $99\%$ and a macro-averaged accuracy of $(100\%+0\%)/2 = 50\%$. The former would in many cases be seen as overoptimistic, while the latter may be seen as overly pessimistic. Which value carries more weight depends on the problem at hand. Therefore, we provide both values for Accuracy.

\textbf{Summary statistics:} In order to capture the performance across multiple datasets in a single number, the benchmark reports the average performance over all datasets $d \in \mathcal{D}$ and metric scores $S= \{\acc, \fone, \jaccard \}$. To identify potential generalization problems, the benchmark also reports the worst case performance across all dataset (i.e. the minimum value for a metric across all datasets). 
\begin{align}
 S_{avg} = \frac{1}{|\mathcal{D}|} \sum_{d \in \mathcal{D}}S_d, & &
 S_{wc} = \min_{d \in \mathcal{D}} S_d
 \end{align}
These summary statistics will become even more important when more datasets become available for inclusion in the benchmark.

\subsection{Benchmark guidelines}

The provided benchmark implementation is intended to facilitate the creation of comparable results across different research groups. There are a couple of benchmark guidelines that have to be followed by each user:

\begin{enumerate}
    \item \textit{All} datasets and \textit{all} metrics should be evaluated to avoid suspicion of cherry-picking.
    \item The intention is to evaluate a single approach on multiple datasets. Therefore, the \textit{same} algorithm or the \textit{same} model architecture should be used for \textit{all} datasets.
    \item The model can be trained individually for each dataset.
    \item If different hyperparameters are used for different datasets, this should be stated explicitly. 
    \item Only train and validation data splits can be used for training and hyperparameter tuning. The test data should only be used once for the final evaluation.
    \item If additional datasets are used for training or parameter tuning or pre-trained models are used, this should be stated explicitly.
    \item We strongly encourage benchmark users to also provide information about the computational effort (runtime, memory, used hardware) and to share implementations and trained models.
\end{enumerate}

\section{Strong baselines}\label{sec:baseline}

We present two strong baselines. One based on a small and well regularized model that uses only the training data provided with the benchmark \gls{hsi} dataset. And a second, larger model that is well suited to leverage additional data through initialization with pre-trained weights.

\subsection{Models}

Our baseline approaches build upon two different model architectures, U-Net \cite{Ronneberger2015UNet} and \gls{dl3} \cite{Chen18DeeplabV3Plus}. These where chosen as they were suitable for our experiments and because they represent well-established architectures for the given task.
The U-Net model implements an Encoder-Decoder-Architecture. In the encoder, the model first compresses data to a discriminative low-resolution feature map and then recovers the original resolution in the decoder, guided by skip-connections in intermediate layers.
In contrast to the original U-Net model we (1) use bilinear upsampling instead of transposed convolutions in the decoder to avoid checkerboard artifacts \cite{Wojna2020Devil}, (2) adapt the channel size of the input layer to allow the training of images with an arbitrary feature dimension, and (3) apply a combination of regularization techniques to account for small training set sizes. The following subsection~\ref{sec:baseline_hyperparams} will demonstrate how the HS3-Bench can be used for hyperparameter tuning of the resulting regularized U-Net (\runet).

The \gls{dl3} model extends the DeeplabV3 \cite{Chen2017RethinkingAC} model. Both models use an \gls{aspp} module to extract rich semantic features at multiple image resolutions without poooling \cite{Chen18DeeplabV3Plus}. To combine this property with the advantage of Encoder-Decoder-based architectures, \gls{dl3} extends DeeplabV3 with a decoder module that guides the model in recovering the original image resolution with intermediate feature maps, allowing it to better recover small objects and object boundaries compared to its predecessor. For initial input processing we use a MobileNetV2 \cite{Sandler2018MobileNetV2IR} backbone network. To train the model with images of arbitrary feature dimension, we introduce a $1 \times 1$ convolutional layer as input layer that reduces the dimensionality to 3 and makes it compatible with the expected input dimension of the original model for RGB data.

\subsection{Using HS3-Bench for Hyperparameter Tuning}\label{sec:baseline_hyperparams}
The proposed HS3-Bench can be used to derive model hyperparameters. As a demonstration, we used this for deriving a regularization configuration for \runet\ in order to reduce the risk of overfitting. 
 We considered \gls{da}, \gls{wd}, \gls{cw}, \gls{bn}, and \gls{do} as potential regularization techniques. We systematically tested different combinations and parameterizations by fitting the model parameters on the HS3-Bench training datasets and evaluated the trained models on the validation set to receive an estimation of the model performance on unseen data. By monitoring the average performance values we were able to identify hyperparameter settings that generalize well across all datasets.
 
 Following this system, we identified \gls{bn} as well as \gls{da} and \gls{do} with respective probabilities of 0.1 and 0.25 as best configuration. \tablename~\ref{tab:hs3baseline} shows the resulting improvement on the HS3-Bench test data. A detailed discussion of the results from this table will follow in section~\ref{sec:experiments}. As an example, compared to regular U-Net the average Jaccard score improves by $+4.32\%$

\subsection{Synthesizing (Pseudo-)RGB Images}\label{subsec:pseudorgb}
It is common ground that in many deep learning applications, access to more data can provide substantial improvements. In driving scenarios, there are large amounts of RGB training images and pre-trained networks available. For example, \gls{dl3} uses a MobileNetV2 backbone network for which model weights pre-trained on ImageNet\footnote{The pre-trained model weights provided by Pytorch were used: \url{https://pytorch.org/vision/stable/models.html}} are publicly available.
To benefit from this additional information we can synthesize RGB images from hyperspectral images. 

We follow the simple approach of Ding et al. \cite{Ding23DualFusion}, by manually selecting three bands from all spectral bands in the \glspl{hsi} that resemble the red, green, and blue channels. We used channels $(63,19,1)$, $(14,7,0)$ and $(2,1,0)$ for \gls{hcv}, \hyko\ and \hsidrive\ respectively. We scale all pixels $\mathbf{p} \in \mathbb{R}^{3}$ in the synthesized image according to the channel-specific min- and max-values  $\mathbf{p}_{min}$, $\mathbf{p}_{max} \in \mathbb{R}^{3}$ derived from the whole dataset. Hence, for each image pixel $\mathbf{p}$ and all images in the dataset we calculate the new pixel values $\mathbf{p'}$
\begin{equation}
    \mathbf{p'} = (\mathbf{p} - \mathbf{p}_{min})/(\mathbf{p}_{max} - \mathbf{p}_{min})
    \label{eq:nomalization}
\end{equation}

\begin{figure}
    \centering
    \includegraphics[width=.8\columnwidth]{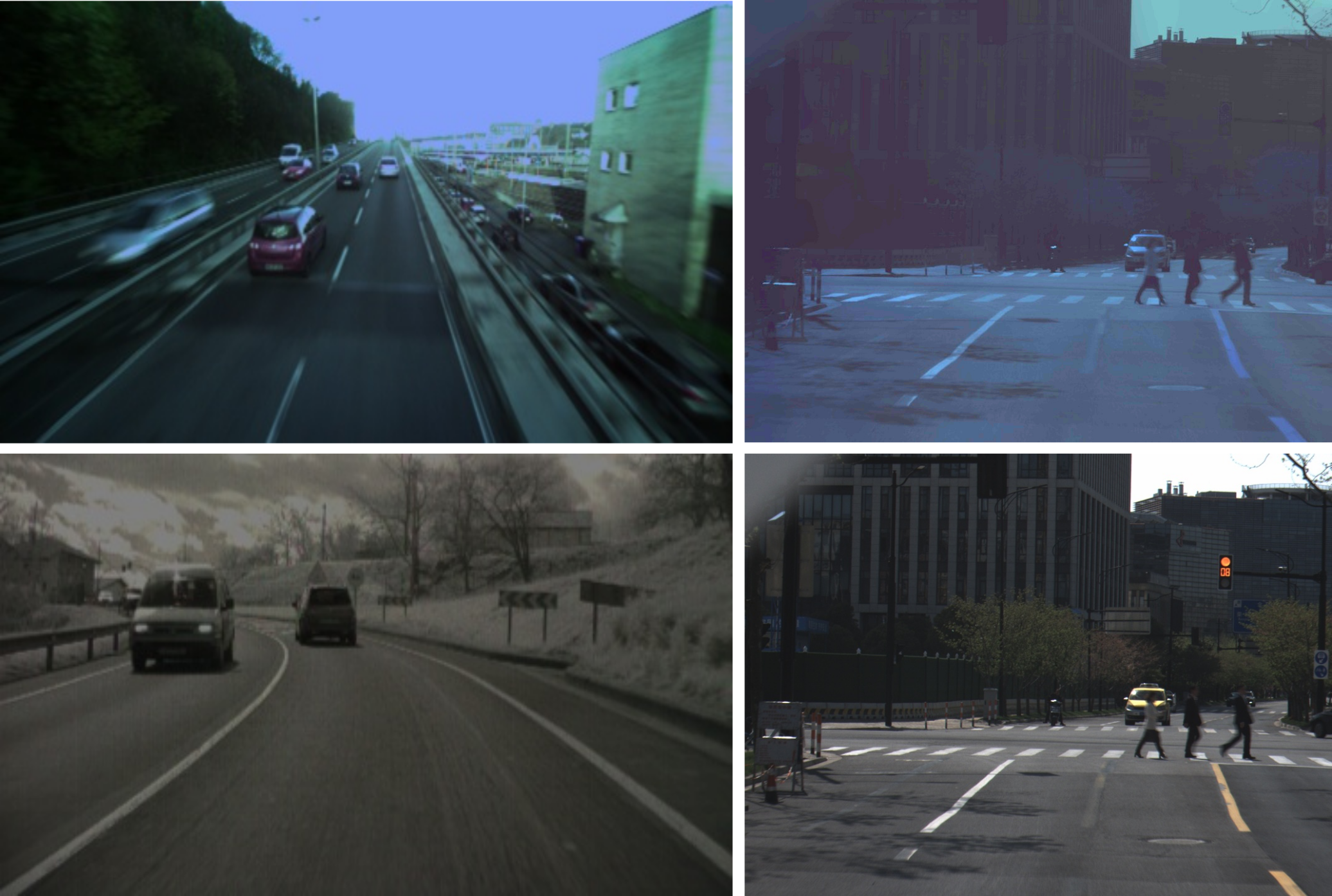}
    \caption{Examples of \gls{prgb} images synthesized from \gls{hsi} images for \hyko\ (top left), \gls{hcv} (top right), \hsidrive\ (bottom left) and RGB images supplied together with \gls{hcv} (bottom right).}\vspace{-0.3cm}
    \label{fig:prgb_examples}
\end{figure}

As \hsidrive\ only covers the red and near-infrared spectrum, which does not allow the estimation of RGB images, we will refer to the synthesized images as \glsfirst{prgb} images from now on. Example images are shown in Fig.~\ref{fig:prgb_examples}. Note that this drastically reduces the data volume and feature dimensionality.

As a special property, \gls{hcv} also provides images from an RGB camera together with the \gls{hsi} data. The RGB images are synchronized and cropped to the same resolution as their \gls{hsi} counterparts. The images were exactly registered such that pixels at the same image coordinates refer to the same object in world coordinates. However, in general such data is not available for \gls{hsi} datasets.
When we use this data for comparison, we will apply the same normalization strategy from eq.~\ref{eq:nomalization}.

\section{Experiments \& Results} \label{sec:experiments}
In this section, we will use HS3-Bench to evaluate the performance of different algorithms and input data configurations for semantic segmentation on \gls{hsi} data.
In section~\ref{sec:hsi}, we first train our baseline models on the \gls{hsi} data and also evaluate the influence of dimensionality reduction on model performance. In section \ref{sec:rgb_comparison} we compare the performance of models using (pseudo-)RGB images to those using full spectrum \gls{hsi}.
Then, we quantify the performance improvement from pre-training on additional data in combination with (pseudo-)RGB data in section~\ref{sec:pretrain}, followed by qualitative assessment of prediction results and a discussion of the improvement of the HS3-baseline models over the state of the art in sections~\ref{sec:qualitative} and \ref{sec:sota}.

Unless stated otherwise, we kept all hyperparameters constant. The parameters are shown in \tablename~ \ref{tab:fixed_training_params}. We used cross-entropy loss for all of our experiments. All experiments were performed on a single Nvidia-A100 GPU with 40GB VRAM.
\begin{table}
    \centering
    \footnotesize
    \caption{Fixed training parameters of all training runs for each dataset, unless otherwise specified for individual runs.}
    \begin{tabular}{lcccc}
        \toprule
        Training Parameter & \hykovis & \gls{hcv}  & \hsidrive  \\\midrule
        optimizer          & AdamW     & AdamW  & AdamW     \\
        learning rate      & $10^{-3}$& $10^{-3}$& $10^{-3}$\\
        optimizer epsilon  & $10^{-8}$ & $10^{-4}$ & $10^{-8}$      \\
        batch size         & \num{16}        & \num{4}     & \num{32}          \\
        max epochs             & \num{500}       & \num{100}   &  \num{300}\\
        early stopping & \checkmark & \checkmark & \checkmark \\
        loss & cross-entropy & cross-entropy & cross-entropy 
        \\\bottomrule
    \end{tabular}\vspace{-0.2cm}
    \label{tab:fixed_training_params}
\end{table}
\tablename~\ref{tab:hs3baseline} provides the main results from this paper. 
The upper part provides results for the individual datasets, the lower part provides summary statistics across all datasets.

\subsection{Applying HS3-Bench for Comparison of Full-Spectrum \gls{hsi} Data and Reduction to a Single Channel}\label{sec:hsi}

We applied the benchmark baseline models U-Net, \runet\ and \gls{dl3} from Sec.~\ref{sec:baseline} on different input data. For example, the entry HSI in the column Data indicates that all spectral bands are used.
As mentioned before, the results in \tablename~\ref{tab:hs3baseline} suggest that \runet\ generally performs better than U-Net.
Notably, this smaller model also performs better than the larger \gls{dl3} model in this comparison where no additional data is available.

\begin{figure}
    \centering
    \begin{minipage}{\columnwidth}
    \includegraphics[width=\columnwidth]{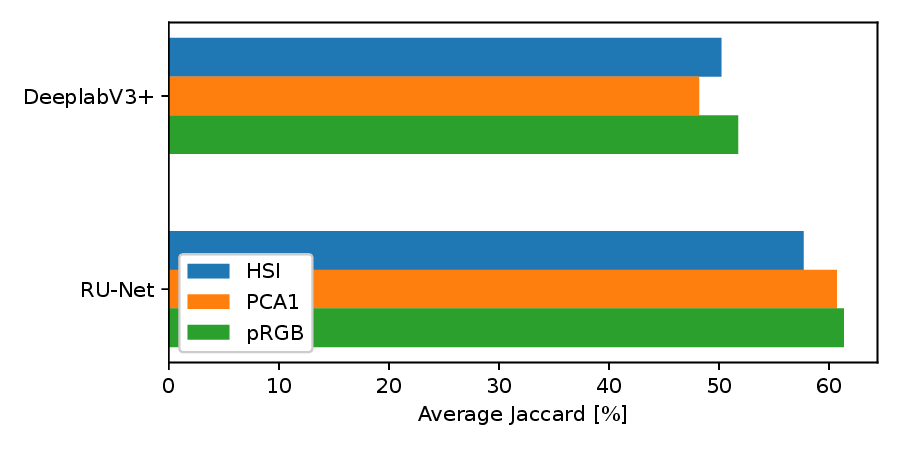}
    \end{minipage}
    \caption{Average Jaccard score per model and data type.}\vspace{-0.3cm}
    \label{fig:avg_jacc}
\end{figure}

Training a model with all spectral bands available in the benchmark datasets defines one extreme, another extreme is reducing the spectral information to a single feature channel with \gls{pca}, denoted as PCA1 in \tablename~\ref{tab:hs3baseline}. Surprisingly, as also depicted in Fig.~\ref{fig:avg_jacc}, the average performance of \gls{dl3} with PCA1 data compared to \gls{hsi} data is only slightly worse. Further, the average performance of \runet\ with PCA1 is even better than with \gls{hsi}. Looking at the individual datasets, we see that on \gls{hcv} both architectures benefit from PCA1. On \hsidrive\ and \hyko, \runet\ profits significantly on the former and has no significant effect on the latter, while the performance of \gls{dl3} is decreased in both cases. We suspect that because \hyko\ has only 16 spectral channels, much fewer than the 128 spectral channels of \gls{hcv}, models trained on the former are less affected by the curse of dimensionality, while models trained on the latter are not presented with enough data to produce a robust classifier.

\subsection{Comparison of \Gls{hsi} and (Pseudo-)RGB Data} \label{sec:rgb_comparison}

In this section we quantify the discrepancy in performance of models trained on \gls{hsi} data and on \gls{prgb} data. 
We synthesized \gls{prgb} images from \gls{hsi} data, as described in section~\ref{sec:baseline}, and used the same dataset splits to keep the class distribution constant in our experiments. 

We repeat the experiments from section~\ref{sec:hsi} with the synthesized \gls{prgb} images. Note that all information in the \gls{prgb} images was derived from \gls{hsi}. We trained our baseline models from scratch and did not use pre-training. The results are also presented in \tablename~\ref{tab:hs3baseline}. Fig.~\ref{fig:avg_jacc} illustrates that the average jaccard score of \runet\ with \gls{prgb} data improves by $+3.66\%$ and for \gls{dl3} by $+1.54\%$ over the performance of the same model trained on \gls{hsi}. The improvement can mainly be traced back to \hyko\ and \gls{hcv}, \hsidrive\ results improve only slightly. \hsidrive\ covers only red and near-infrared channels so we could not select bands from the spectral intervals corresponding to RGB-wavelengths. Therefore, the results might be explained by suboptimal band selection during RGB image synthesis. In summary using \gls{prgb} data leads to an overall improvement in model performance.

\begin{table}
    \centering
    \footnotesize
    \caption{Benchmark scores (\%) on the HS3-Bench test data.}.    \hspace*{-0.7cm}
    %\scriptsize
    \resizebox{\columnwidth}{!}{%
    \begin{tabular}{llr|cccc}
                    \toprule
                        
                       &&& \multicolumn{4}{c}{Testing}  \\\cmidrule(lr){4-7}
                       Dataset & Approach & Data & $\acc_\micro$ & $\acc_\macro$ & $\foneM$ & $\jaccard_\macro$
                       \\\midrule
                       \gls{hcv} & U-Net & \gls{hsi} & 85.25 & 48.62 & 48.18 & 37.73\\
                                 & \runet & \gls{hsi} & 87.63 & 54.14 & 53.26 & 42.23\\
                                 & \runet & PCA1 & 88.25 & 58.07 & 55.43 & 44.26\\
                                 & \runet & \gls{prgb} & 87.95 & 56.65 & 55.46 & 44.03 \\
                                 & \gls{dl3} & \gls{hsi} & 86.60 & 53.15 & 51.83 & 40.79 \\
                                 & \gls{dl3} & PCA1 & 86.64 & 54.46 & 52.90 & 41.58\\
                                 & \gls{dl3} & \gls{prgb} & 87.00 & 55.33 & 54.08 & 42.58\\
                                 & \gls{dl3}(BB) & \gls{prgb} & \textbf{90.26} & \textbf{64.10} & \textbf{61.93} & \textbf{50.04}\\
                                 & (\gls{dl3}(BB))\tablefootnote{The result on RGB data is surrounded in parenthesis as the data was collected with an additional sensor (cf. Sec.~\ref{sec:rgb_comparison}). RGB data was only provided for \gls{hcv} and therefore we could not calculate summary statistics.} & (RGB) & (91.22) & (65.87) & (63.33) & (52.11)\\
                                 & \gls{dl3}(PT) & \gls{prgb} & 89.62 & 61.91 & 60.17 & 48.47\\
                                 \midrule
                       \hyko & U-Net & \gls{hsi} & 85.36 & 68.15 & 68.55 & 57.39\\
                                 & \runet & \gls{hsi} & 86.72 & 68.79 & 69.19 & 58.64\\
                                 & \runet & PCA1 & 85.61 & 68.09 & 70.01 & 58.67 \\
                                 & \runet & \gls{prgb} & 89.18 & 73.92 & 75.04 & 64.67\\
                                 & \gls{dl3} & \gls{hsi} & 84.10 & 63.01 & 64.90 & 53.22\\
                                 & \gls{dl3} & PCA1 & 79.99 & 61.59 & 63.00 & 50.40\\
                                 & \gls{dl3} & \gls{prgb} & 84.64 & 65.30 & 66.56 & 54.82\\      
                                 & \gls{dl3}(BB) & \gls{prgb} & \textbf{90.49} & \textbf{74.87} & \textbf{77.11} & \textbf{66.77}\\
                                 & \gls{dl3}(PT) & \gls{prgb} & 88.62 & 73.97 & 76.79 & 65.41\\
                                \midrule
                       \hsidrive & U-Net & \gls{hsi} & 94.95 & 74.74 & 76.08 & 64.95 \\
                                 & \runet & \gls{hsi} & 96.08 & 79.82 & 82.34 & 72.18\\
                                 & \runet & PCA1 & 97.02 & \textbf{86.80} & \textbf{87.76} & \textbf{79.23}\\
                                 & \runet & \gls{prgb} & 96.32 & 82.70 & 84.91 & 75.31 \\
                                 & \gls{dl3} & \gls{hsi} & 92.51 & 65.58 & 67.86 & 56.63 \\
                                 & \gls{dl3} & PCA1 & 90.88 & 62.93 & 64.31 & 52.62\\
                                 & \gls{dl3} & \gls{prgb} & 92.74 & 66.59 & 69.46 & 57.84\\
                                 & \gls{dl3}(BB) & \gls{prgb} & \textbf{97.09} & 83.93 & 86.41 & 77.44\\
                                 & \gls{dl3}(PT) & \gls{prgb} & 95.69 & 81.95 & 84.09 & 73.84\\
                                 \midrule \midrule
                       \textbf{Average}     & U-Net & \gls{hsi} & 88.52 & 63.84 & 64.27 & 53.36\\
                       \textbf{Perf.} & \runet & \gls{hsi} & 90.14 & 67.58 & 68.26 & 57.68 \\
                                        & \runet & PCA1 & 90.29 & 70.99 & 71.07 & 60.72\\
                                        & \runet & \gls{prgb} & 91.15 & 71.09 & 71.80 & 61.34 \\
                                      & \gls{dl3} & \gls{hsi} & 87.74 & 60.58 & 61.53 & 50.21 \\
                                      & \gls{dl3} & PCA1 & 85.84 & 59.66 & 60.07 & 48.20 \\
                                      & \gls{dl3} & \gls{prgb} & 88.13 & 62.41 & 63.37 & 51.75 \\
                                 & \gls{dl3}(BB) & \gls{prgb} & \textbf{92.61} & \textbf{74.30} & \textbf{75.15} & \textbf{64.75}\\
                                 & \gls{dl3}(PT) & \gls{prgb} & 91.31 & 72.61 & 73.68 & 62.57\\

                       \midrule
                       \textbf{Worst-Case}  & U-Net & \gls{hsi} & 82.25 & 48.63 & 48.18 & 37.73\\
                       \textbf{Perf.} & \runet & \gls{hsi} & 86.72 & 54.14 & 53.26 & 42.23 \\
                                      & \runet & PCA1 & 85.61 & 58.07 & 55.43 & 44.26 \\
                                      & \runet & \gls{prgb} & 87.95 & 56.65 & 55.46 & 44.03 \\
                                        & \gls{dl3} & \gls{hsi} & 84.10 & 53.15 & 51.83 & 40.79 \\
                                        & \gls{dl3} & PCA1 & 79.99 & 54.46 & 52.90 & 41.58 \\
                                        & \gls{dl3} & \gls{prgb} & 84.64 & 55.33 & 54.08 & 42.58 \\
                                 & \gls{dl3}BB & \gls{prgb} & \textbf{90.26} & \textbf{64.10} & \textbf{61.93} & \textbf{50.04}\\
                                 & \gls{dl3}PT & \gls{prgb} & 88.62 & 61.91 & 60.17 & 48.47\\
                       \midrule
                       \bottomrule
    \end{tabular}
    }\vspace{-0.2cm}
    \label{tab:hs3baseline}
\end{table}

\subsection{Impact of Pretraining on Model Performance}\label{sec:pretrain}

To investigate the potential benefit of pre-training on RGB data we compare the performance of the \gls{dl3} model on \gls{prgb} data without pre-training and with pre-training. In our first test, we intialize the backbone networks of \gls{dl3} with pre-trained weights (cf. \ref{sec:baseline}) and fine-tune the full model on our data. Then, to see how well information from similar domains can be transferred we initialize \textit{all} of our model parameters with model weights pre-trained on Cityscapes\footnote{The model weights were downloaded from this repository: \url{https://github.com/VainF/DeepLabV3Plus-Pytorch}} \cite{Cordts2016TCD}. We only replaced the output layer, such that the number of predicted classes matched the number of classes of each dataset in HS3-Bench. In our second test \textit{all} model parameters were frozen, except the ones in the output layer to avoid adapting the models feature extraction to the new dataset.  

\begin{figure}
    \centering
    \begin{minipage}{\columnwidth}
    \includegraphics[width=\columnwidth]{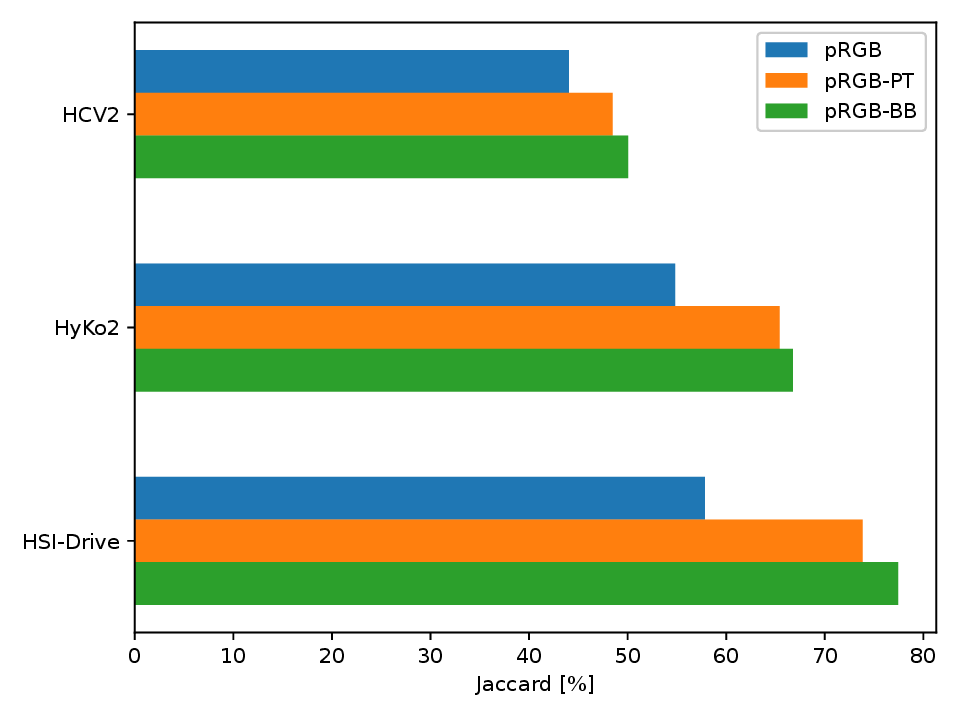}
    \end{minipage}
    \caption{Jaccard scores for \gls{dl3} on \gls{prgb} data per dataset without pre-training (\gls{prgb}), with pre-trained backbone and fine-tuning (\gls{prgb}-BB) and with pre-trained weights transferred from a similar domain, namely CityScapes (\gls{prgb}-PT). In the latter, all weights except for output-layer were frozen.}\vspace{-0.3cm}
    \label{fig:pretrained}
\end{figure}

The result of using pre-trained weights and fine-tuning the full model are summarized in \tablename~\ref{tab:hs3baseline} denoted with BB. The average performance increased significantly, by around $+7\%$ compared to the best model that does not use pre-trained weights. Fig.~\ref{fig:pretrained} shows that the performance improvement compared to \gls{prgb} images without pre-training is most apparent in \hsidrive. This is especially interesting, as \hsidrive\ contains only spectral information from the red and the near-infrared spectrum. Hence, the synthesized images do not have the same distribution typically apparent in RGB images. This observeration indicates that the features extracted by the backbone model are general enough to be successfully applied to neighboring spectral domains. 

In our second experiment on pre-training, fine-tuning only the output layer showed only slightly lower average performance than fine-tuning the full model. Nonetheless, the performance is still significantly better than all approaches that do not use pre-training, which shows that the feature extraction modules can be directly applied to similar domains. 
When only considering \gls{prgb} data all results on each individual dataset improve with pre-training by a large margin, as depicted in Fig.~\ref{fig:pretrained}. 
The strong improvements in model performance indicate that exploiting knowledge through pre-training is very effective. It has an even stronger relative effect than using \gls{prgb} data instead of full-spectrum \gls{hsi}.
Further, the observation that models using \gls{prgb} data outperformed all models using full spectra, support that for driving scenarios introducing knowledge from related domains is more beneficial than adding additional spectral features for the available datasets.

Note that we used a very simple way of synthesizing \gls{prgb} images (see Sec. \ref{subsec:pseudorgb}). The spectral bands are very narrow - especially in \gls{hcv} - which leads to lower signal intensities and in turn to noisy bands. The synthesized images show an unnatural color distribution (see Fig.~\ref{fig:prgb_examples}), distinct from typical RGB-images. 
To estimate the upper limit of model performance that can be expected with better RGB-image synthesis, we trained an additional model on the RGB images provided with \gls{hcv} (cf. section~\ref{sec:rgb_comparison}). We fine-tuned \gls{dl3} using hyperparameter settings from \tablename~\ref{tab:fixed_training_params} and a pre-trained MobileNetV2 backbone and achieved a Jaccard score of $52.11\%$ ($+0.43\%$ as compared to the current state-of-the-art results that were published in \cite{Ding23DualFusion}). Hence, we expect that with more sophisticated RGB image synthesis methods the results on all data sets are likely to improve.

\subsection{Qualitative Evaluation} \label{sec:qualitative}
\begin{figure}
    \centering
    \begin{minipage}{0.25\columnwidth}
    \includegraphics[width=\columnwidth]{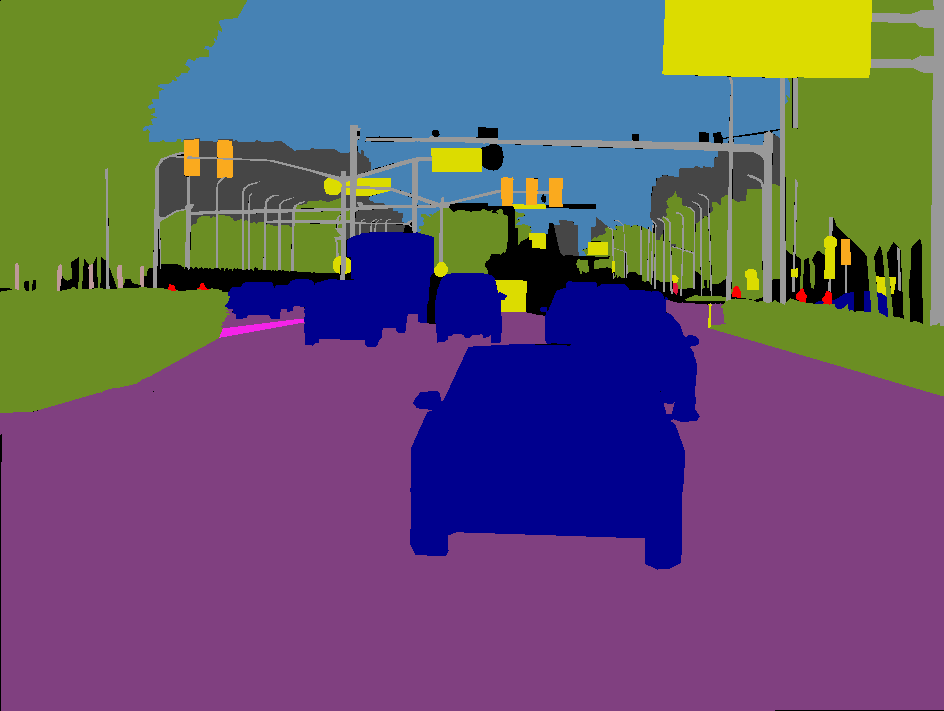}
    \end{minipage}
        \begin{minipage}{0.32\columnwidth}
    \includegraphics[width=\columnwidth]{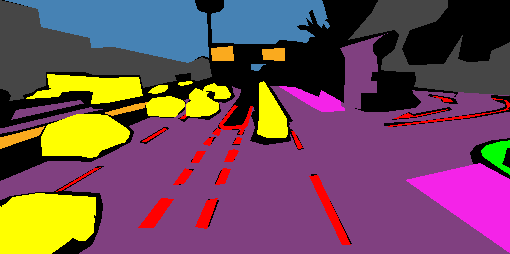}
    \end{minipage}
        \begin{minipage}{0.32\columnwidth}
    \includegraphics[width=\columnwidth]{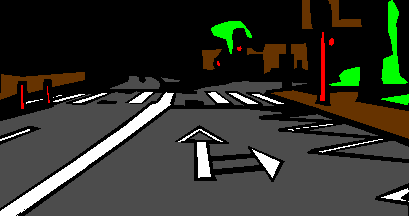}
    \end{minipage}\\[.2em]
        \begin{minipage}{0.25\columnwidth}
    \includegraphics[width=\columnwidth]{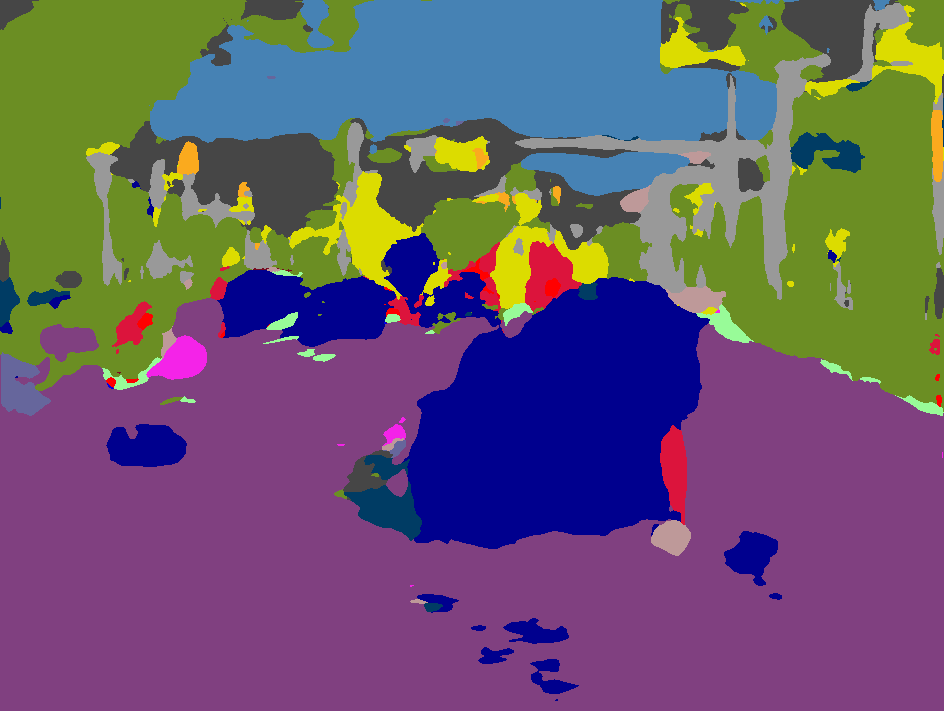}
    \end{minipage}
        \begin{minipage}{0.32\columnwidth}
    \includegraphics[width=\columnwidth]{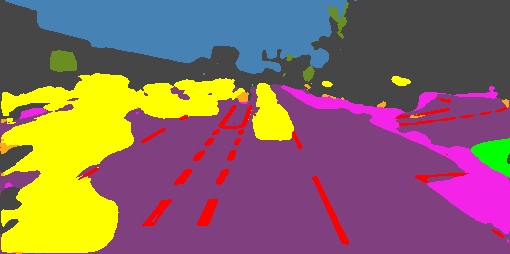}
    \end{minipage}
        \begin{minipage}{0.32\columnwidth}
    \includegraphics[width=\columnwidth]{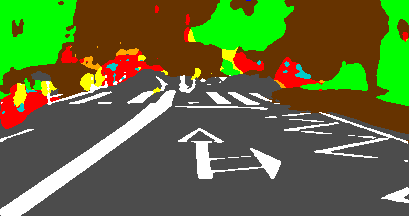}
    \end{minipage}\\[.2em]
        \begin{minipage}{0.25\columnwidth}
    \includegraphics[width=\columnwidth]{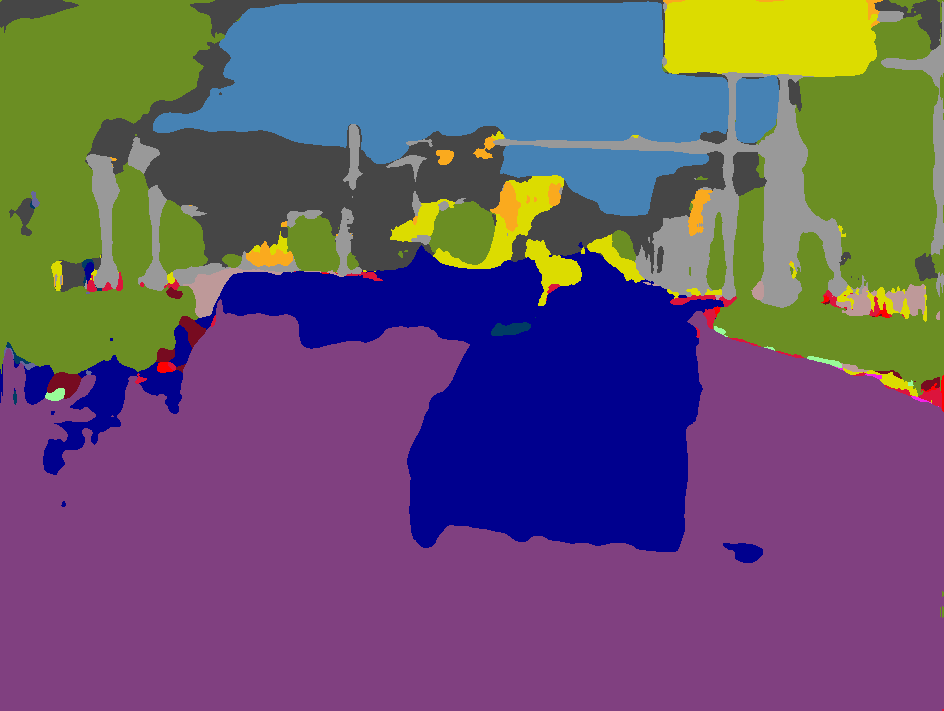}
    \end{minipage}
        \begin{minipage}{0.32\columnwidth}
    \includegraphics[width=\columnwidth]{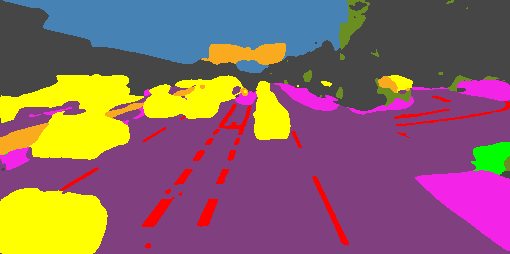}
    \end{minipage}
        \begin{minipage}{0.32\columnwidth}
    \includegraphics[width=\columnwidth]{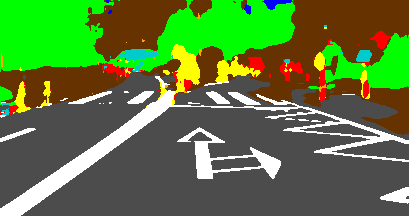}
    \end{minipage}\\[.2em]
        \begin{minipage}{0.25\columnwidth}
    \includegraphics[width=\columnwidth]{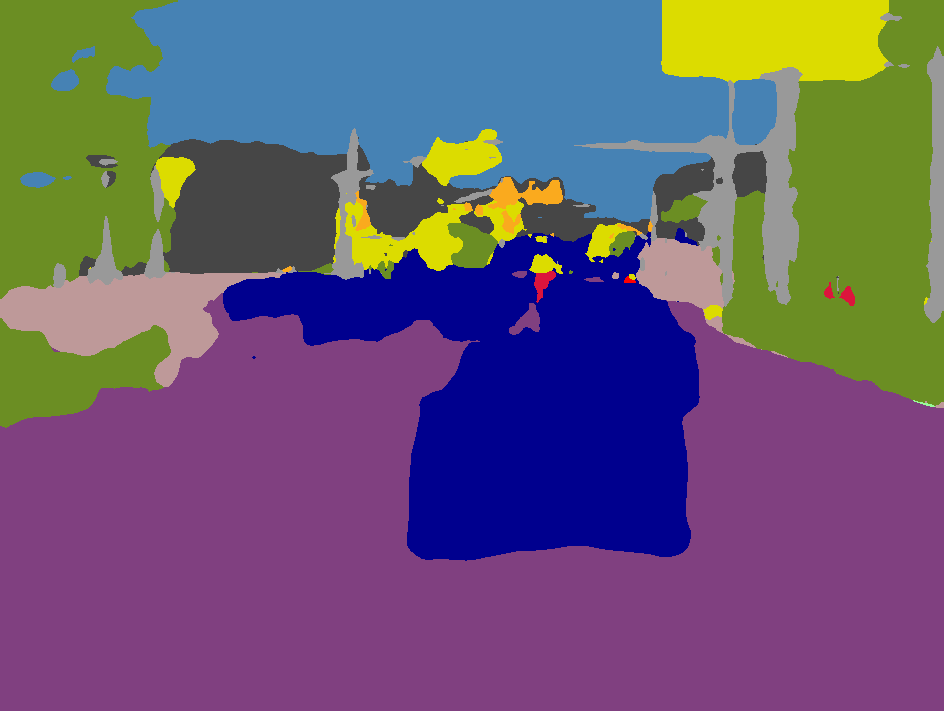}
    \end{minipage}
        \begin{minipage}{0.32\columnwidth}
    \includegraphics[width=\columnwidth]{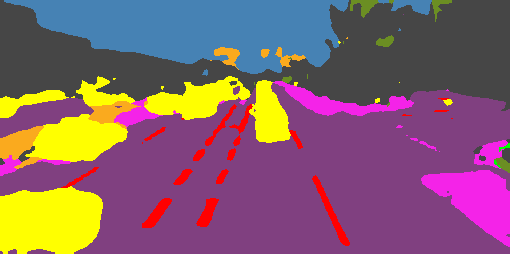}
    \end{minipage}
        \begin{minipage}{0.32\columnwidth}
    \includegraphics[width=\columnwidth]{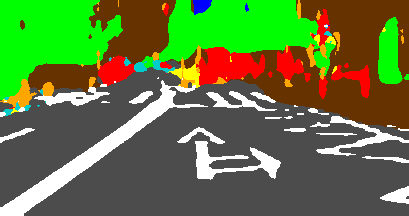}
    \end{minipage}
    \caption{Example inferences for \gls{hcv}, \hyko\ and \hsidrive. The top row shows the ground-truth label map. Row 2 and 3 show inferences on \gls{hsi} and \gls{prgb} data without pre-training, respectively.
    The bottom row shows inferences with \gls{prgb} data with pre-training. We consistently used the models that showed best performance for the given data type, \ie\ \runet (row 2 and 3) and \gls{dl3} with pretrained backbone (row 4).}
    \label{fig:visual_comp}
\end{figure}

To give a visual impression of the models segmentation performance, Fig.~\ref{fig:visual_comp} shows example inferences for each data set in HS3-Bench side by side. The top row shows ground-truth label maps, followed by inferences on \gls{hsi} data, then \gls{prgb} and finally \gls{prgb} with pre-trained backbone. For inference we applied the model that showed the best average performance for the given data type or pre-training configuration consistently to all datasets. The best models are \runet\ for \gls{hsi} and \gls{prgb} without pre-training (row 2 and 3) and fine-tuned \gls{dl3} with pre-trained backbone network (row 4). The example predictions support the impression of the statistical results. The predictions on \gls{prgb} data are less noisy than \gls{hsi} and contours are more precise. The difference between row 3 and row 4 are subtle. It seems that object contours are a bit more precise for the pre-trained model and object surfaces are more homogeneous.
 
\subsection{Comparison to the State of the Art} \label{sec:sota}
\gls{hcv} was introduced in the context of a challenge for a workshop at ICCV 2021. The best reported results in the competition achieved a Jaccard score of $51.4\%$. In \cite{Ding23DualFusion} this result was raised to $51.76\%$ by HRNet \cite{Wang21HRNet} that was fine-tuned on the RGB-images provided with \gls{hcv}. Our best pre-trained model on these RGB-images achieved a Jaccard score of $52.11\%$ ($+0.43\%$ increase compared to the current state-of-the-art).

Under the conditions, that no pre-training and only \gls{hsi} data or data derived from \gls{hsi} is used, the best listed model (FCN101 \cite{Long15FCN}) in \cite{Ding23DualFusion} achieved a Jaccard score of $41.13\%$. With \runet\ and \gls{hsi} data we improved this result to $42.23\%$ ($+1.1\%$) and with data derived from \gls{hsi}, \ie\ PCA1, the same model further improves to $44.26\%$ ($+3.13\%$).

\section{Summary} \label{sec:summary}
In this paper we presented HS3-Bench, a hyperspectral semantic segmentation benchmark for driving scenarios which is designed for systematic comparison of different models and algorithms. Based on this benchmark, we performed systematic evaluation of hyperspectral image representations, \ie\ full spectrum, \gls{pca}-reduced spectrum and synthesized pseudo-RGB images as well as the impact of knowledge transfer through pre-trained weights. 
We demonstrated the application of HS3-Bench by deriving a suitable configuration of regularization approaches to a U-Net model (\runet). In our experiments we used \runet\ as well as \glsfirst{dl3} with a MobileNetV2 backbone. 

We consider both models - \runet\ and \gls{dl3} -  as strong baselines for HS3-Bench. Under the condition that only limited hyperspectral data is available the regularized U-Net with dimensionality reduction outperforms \gls{dl3} as well as the current state of the art model. However, if additional RGB data is available in the problem domain, \gls{dl3} with \gls{prgb} images synthesized from \gls{hsi} data can effectively leverage the domain knowledge through pre-training and should be preferred. \gls{dl3} with a pre-trained backbone network fine-tuned on RGB data outperforms the previous state-of-the-art models using pre-trained weights as well. 

Our results pose interesting questions for future research. 
In \cite{Ding23DualFusion} the authors state that their dual fusion network effectively utilizes knowledge from pre-trained RGB models and hyperspectral data. However, our results suggest that major improvements can be traced back to leveraging domain knowledge through pre-trained model parameters. Further, our experiments support that available learning-based models benefit more from leveraging additional RGB training data than from leveraging additional HSI channels. 
We believe the proposed HS3-Bench can be a valuable tool to support research directions such as finding general backbone models for \gls{hsi} data and models that better exploit all channel information in \gls{hsi} data. Also, further investigation is required to identify the causes of the performance discrepancy between \gls{hsi} and RGB.

%%%%%%%%%%%%%%%%%%%%%%%%%%%%%%%%%%%%%%%%%%%%%%%%%%%%%%%%%%%%%%%%%%%%%%%%%%%%%%%%

\bibliographystyle{IEEEtran}
\bibliography{IEEEabrv,refs}

\end{document}